\documentclass[conference]{IEEEtran}
\IEEEoverridecommandlockouts
\usepackage{cite}
\usepackage{amsmath,amssymb,amsfonts}
\usepackage{algorithmic}
\usepackage{graphicx}
\usepackage{textcomp}
\usepackage{hyperref}
\usepackage{booktabs}
\usepackage{adjustbox}
\usepackage{diagbox}
\usepackage{calc}
\usepackage{float}
\usepackage{dblfloatfix}
\usepackage{url}
\usepackage{svg}
\usepackage{xcolor}
\usepackage{paralist}
\usepackage{hyperref}

\usepackage{orcidlink}

\usepackage{graphicx}
\usepackage{subcaption}

\def\svgwidth{500 pt}

\def\BibTeX{{\rm B\kern-.05em{\sc i\kern-.025em b}\kern-.08em
    T\kern-.1667em\lower.7ex\hbox{E}\kern-.125emX}}

\newif\ifdraft

\ifdraft
\newcommand{\note}[1]{ {\textcolor{orange} { **Note: #1 }}}
\newcommand{\alnote}[1]{ {\textcolor{blue} { ***Andre: #1 }}}
\newcommand{\jpnote}[1]{ {\textcolor{green} { ***Josef: #1 }}}
\newcommand{\prnote}[1]{ {\textcolor{orange} { ***Philipp: #1 }}}
\newcommand{\crinoteone}[1]{ {\textcolor{red} { ***Reviewer 1: #1 }}}
\newcommand{\critnotetwo}[1]{ {\textcolor{yellow} { ***Reviewer 2: #1 }}}
\newcommand{\critnotethree}[1]{ {\textcolor{purple} { ***Reviewer 3: #1 }}}
\newcommand{\critnotefour}[1]{ {\textcolor{magenta} { ***Reviewer 4: #1 }}}
\newcommand{\critnotefive}[1]{ {\textcolor{orange} { ***Reviewer 5: #1 }}}
\else
\newcommand{\note}[1]{}
\newcommand{\alnote}[1]{}
\newcommand{\prnote}[1]{}
\newcommand{\jpnote}[1]{}
\newcommand{\crinoteone}[1]{}
\newcommand{\critnotetwo}[1]{}
\newcommand{\critnotethree}[1]{}
\newcommand{\critnotefour}[1]{}
\newcommand{\critnotefive}[1]{}
\fi
    
\begin{document}

\title{Performance Characterization of Expert Router for Scalable LLM Inference}
\author{
    \IEEEauthorblockN{Josef Pichlmeier}
    \IEEEauthorblockA{
        \textit{BMW Group}\\
        \textit{Ludwig Maximilians Universität}\\
        Munich, Germany\\
        Josef.Pichlmeier@bmw.de
    }
    \and
    \IEEEauthorblockN{Philipp Ross\,\orcidlink{0000-0002-4720-9835}}
    \IEEEauthorblockA{
        \textit{BMW Group}\\
        Munich, Germany
    }
    \and
    \IEEEauthorblockN{Andre Luckow\,\orcidlink{0000-0002-1225-4062}}
    \IEEEauthorblockA{
        \textit{BMW Group}\\
        \textit{Ludwig Maximilians Universität}\\
        Munich, Germany
    }
}

\maketitle
\begin{abstract}
Large Language Models (LLMs) have experienced widespread adoption across scientific and industrial domains due to their versatility and utility for diverse tasks. Nevertheless, deploying and serving these models at scale with optimal throughput and latency remains a significant challenge, primarily because of LLMs' high computational and memory demands. Specialized models optimized for specific tasks can be combined through a routing mechanism to address these challenges, creating a modular inference system. This paper introduces Expert Router, a scalable routing architecture that directs prompts to specialized expert models. We characterize multiple Expert Router configurations, including different LLama 3 models with quantized and non-quantized weights under up to 1,000 concurrent users. Our findings reveal that Expert Router introduces minimal latency overhead, with the configuration of expert models being a dominating factor in performance outcomes. High-parameter expert models deliver stable throughput and latency under moderate concurrency levels. In contrast, smaller expert models maintain competitive performance across a wider range of concurrent users compared to tensor-parallelized baseline models. This highlights the potential of Expert Router for efficient and scalable LLM deployment.

\end{abstract}


\section{Introduction}

\begin{figure*}[!ht]
    \begin{center}
    \def\svgwidth{500 pt}
    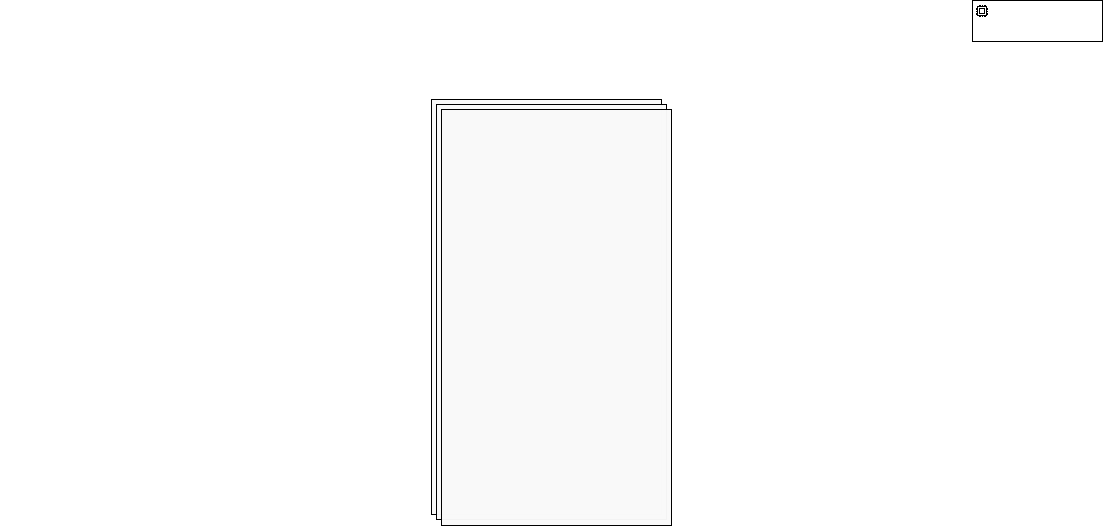
    \caption{\textbf{Expert router architecture and experimental setup:} The architecture comprises three main components: the routing gateway, the Triton inference server, and the user simulator. Incoming prompts (1) are classified by the routing gateway using a k-means algorithm (2 + 3) and forwarded to the corresponding language model (4). All models run on different GPUs independently and can process queries concurrently. The inference responses are returned to the respective users (5 + 6). The clustering algorithm has been trained on a set of samples indicated by the dots, and new prompts are classified according to the respective topics indicated by the colors. }
    \label{fig: Architecture}
    \end{center}
\end{figure*}

Foundation models, particularly large language models (LLMs) like GPT-4~\cite{openai2024gpt4}, Claude 3~\cite{anthropic2024claude}, and Llama 3~\cite{dubey2024llama3herdmodels} offer versatile applications across diverse industrial and scientific domains~\cite{bommasani2022opportunities}, such as the automotive domain~\cite{rony2023carexpert}, software development~\cite{alshahwan2024assured}, physics~\cite{gruver2023finetuned}, and medicine~\cite{Clusmann2023}. These models are based on the transformer architecture~\cite{Vaswani2017}, which uses the self-attention mechanism to calculate relationships within the input sequence and can solve many natural language processing problems, including sentiment analysis \cite{Liang2022} and classification~\cite{zhang_pushing_2024}. More recently, generative tasks such as text summarization, text generation, and conversational systems have become important. Many of these tasks demand multiple LLM inferences, e.\,g., to execute complex prompt sequences derived from different prompt engineering approaches~\cite{yao2023tree}, to utilize data integrations via retrieval augmentation generation, or to orchestrate agent-based workflows for optimal results~\cite{wu2023autogen}.

The need to efficiently conduct inference and serve these models is critical to enable these applications. However, the deployment of LLMs presents significant challenges. This includes the complex model architecture and addressing their high computational and memory demands, often necessitating specialized hardware like GPUs. For larger models, e.\,g., models beyond 70 billion parameters, it is required to partition the model across multiple GPUs utilizing model parallelism (e.\,g. tensor and pipeline parallelism) or other optimization methods (e.\,g. pruning and quantization). 

A different approach to larger models is the utilization of expert models that are tailored to perform well in specific tasks. Combining multiple specialized models through a routing mechanism, which directs queries to the most suitable LLM and creates a single inference system that improves overall answer quality. Recent studies have shown that routing-based inference systems, when configured with smaller LLMs, can outperform larger models regarding response quality \cite{shnitzer_large_2023}. 

However, beyond improvements in response quality, routing inference systems also impact critical performance characteristics such as latency and throughput. Thereby, the routing architecture should maintain stable latency and throughput, even when operating under the load of multiple concurrent users. Furthermore, these systems should be capable of working with different model configurations, enabling them to utilize specific model strengths to meet different performance requirements. Additionally, routing systems should be scalable and adaptable, functioning efficiently in different computing environments. 

In this work, we introduce \emph{Expert Router}, a scalable routing inference architecture that combines multiple expert models into one system. Expert Router utilizes a technique called c-BTM introduced by Gururangan et\,al.~\cite{Gururangan2023}. It employs unsupervised clustering using k-means to partition a given dataset into distinct domains as the basis for multiple expert models trained in parallel. We adopt this architecture to create an embarrassingly parallel inference pipeline where incoming prompts are routed to the correct expert model. To evaluate Expert Router, we conduct extensive experiments with up to 1,000 concurrent users, recording the timestamp of each output token to generate the relevant system and user-centric metrics. This data provides insights into the characteristics of Expert Router, allowing us to assess the impact of routing on latency and throughput. We summarize our contributions in this work as follows: 
\begin{compactitem}
    \item We present Expert Router, a flexible and scalable architecture that efficiently combines multiple expert models, enabling dynamic routing and seamless integration of various model configurations to optimize LLM inference.
    \item We evaluate multiple configurations of Expert Router, including quantized and non-quantized LLMs, benchmarking them against baseline models across user and system metrics. 
    \item Based on our findings, we provide recommendations for optimizing routing-based inference systems, focusing on balancing latency, throughput, and resource allocation.
\end{compactitem}
\section{Related Work}

Multiple studies~\cite{aggarwal_automix_2024, yue_large_2024, chen_frugalgpt_2023, ong_routellm_2024, nguyen_metallm_2024} have explored LLM routing strategies to optimize inference costs. These studies route simple requests to weaker LLMs and direct more complex tasks to more capable models, such as GPT-4. For instance, Wang et al.~\cite{Wang2023} proposed an inference system that uses dynamic routing techniques to distribute queries among multiple small models and complex LLMs. This approach uses less resource-intensive models for simpler queries rather than relying on an LLM for every task.

Other studies \cite{lu_routing_2023, srivatsa_harnessing_2024, yu_breaking_2024-1} have investigated training a routing function to direct prompts to the most suitable LLM based on their capabilities. As an example, Shnitzer et\,al. \cite{shnitzer_large_2023} are using scoring or ranking models to evaluate the strengths and weaknesses of different LLM candidates. The study identifies that it is possible to outperform a larger model by routing to smaller LLMs.

In the field of LLM performance evaluation, recent studies have established metrics that allow benchmarking and comparing different deployment methods. Most notably, Reddie et\,al. \cite{reddi_mlperf_2020} and MLPerf have released a benchmark framework and reference implementations to evaluate inference backends based on a LLama 2 70B model \cite{atta-fosu_llama_2024}. The group defines and describes metrics such as the Time to First Token (TTFT), time per output token (TPOT) and throughput in token per second (token/s). These metrics are also used by other publications to compare the performance of hardware and deployment architectures \cite{agrawal_taming_2024, zhou_survey_2024, zhong_distserve_2024, agrawal_vidur_2024}. Additionally to the above streaming related quantities, multiple studies \cite{Orca, kundu_performance_2024-1, stojkovic_dynamollm_2024} use classical serving metrics such as the total latency or the p99 latency.

LLM architectures can undergo different runtime conditions depending on factors such as the available hardware, number of requests and deployment hyperparameters. To test these different settings it is necessary to measure the change of the mentioned metrics under varying load. Previous work has investigated how factors such as the number of input tokens, batch size and number of concurrent users can affect these metrics \cite{stojkovic_dynamollm_2024, NEURIPS2023_ce7ff340, patel_splitwise_2023}. In the aforementioned MLPerf benchmark, the simulated queries cover a broad spectrum of input token counts, which reflects varying load scenarios \cite{atta-fosu_llama_2024}. Agrawal et\,al. introduced Viduar: A Large-Scale Simulation Framweork for LLM Inference \cite{agrawal_vidur_2024}. In their framework, different workloads can be selected which are based on different input datasets, request types and arrival rates. 

Our work explores the effects of model routing on user and system metrics, particularly under a high number of concurrent users. While previous research has shown that routing can improve answer quality, our study focuses on how it influences different latency metrics and throughput values.
\section{System Architecture}
\label{sec:methods}

This section presents Expert Router's architecture, which enables efficient and scalable LLM inference. The architecture is highly modular, i.\,e., individual components can be replaced or upgraded, allowing for continuous and systematic improvements. For example, the architecture enables LLMs to be individually replaced without necessitating a complete system update. Expert Router can be deployed in a heterogeneous computing environment irrespective of the hardware components' manufacturing era. 

Figure~\ref{fig: Architecture} illustrates the system architecture. The routing gateway is the central component of the system.  It manages request distribution to the inference backend. Specifically, we utilize NVIDIA Triton~\cite{Triton} and TensorRT-LLM~\cite{tensorrtLLM} to execute the inference tasks. The system can easily be scaled on every layer, e.g., utilizing a load balancer like NGNIX before the routing gateway or adding new Triton server instances. It utilizes the gRPC protocol~\cite{gRPC2023} for communication between these components. 

\subsection{Routing Gateway}
\label{subsec: routing}

The routing gateway acts as an intermediary; it receives and distributes requests across inference servers. This functionality is also referred to as reverse proxy. After receiving a new inference request, the routing gateway extracts the user prompt. It sends it to the query classification (Step 3/4 in Figure~\ref{fig: Architecture}), where a k-means clustering algorithm is used to determine the target cluster of the prompt. This mechanism is adopted from the system developed by Gururangan et\,al.~\cite{Gururangan2023}. 

We follow the original paper's preprocessing steps: excluding stop-words and substituting numerical tokens. The clustering process transforms the prompt text data using a TF-IDF vectorizer from the scikit-learn library~\cite{scikit-learn}. The TF-IDF vectors are then condensed to 100 dimensions via a singular value decomposition. This dimensionality reduction is followed by a normalization step by subtracting the mean and scaling the vectors to unit variance \cite{Gururangan2023}. We employ the custom PyTorch implementation, including the weights of the k-means model published by~\cite{Gururangan2023}, which has been trained on a selected portion of the C4 dataset \cite{raffel_exploring_2023}. Working with the reduced-dimensional input sequence is especially efficient in inference scenarious, as the classification process only scales with the number of concurrent users but not with the length of the input requests.

Expert Router uses the pre-trained k-means model to classify the prompt into an appropriate cluster when receiving a new prompt. The routing gateway then directs the prompt to the designated language model hosted on an inference backend instance. Unlike c-BTM, which dynamically selects and combines multiple experts during inference, our approach streamlines the process by routing each request directly to a single, most relevant model, reducing computational overhead during real-time inference. To increase prompt classification throughput, we are deploying 16 routing gateway instances in combination with an NGINX load balancer that distributes the incoming requests in a round robin mechanism~\cite{nginx} (Step 1/2 and 7/8 in Figure~\ref{fig: Architecture}). 

\subsection{Inference Backend and Triton Server}
\label{subsec: triton}

Following an introduction into LLM inference techniques, we discuss the particular inference backend employed by Expert Router: the GPU-accelerated NVIDIA Triton server.

\paragraph*{LLM Inference}
We rely on advanced inference techniques, particularly optimized for the self-attention mechanisms and batching. Self-attention enables every token of the input sequence to be evaluated against all preceding tokens, thereby creating a contextual representation. This process starts by converting input tokens into Query (Q), Key (K), and Value (V) vectors through trained weights. Next, it calculates the dot products between all Q and K vectors and applies a softmax function.  The outcome of this calculation is combined with the V vectors through another dot product to produce a weighted sum. This weighted sum models the context between tokens. A critical optimization within this process is the use of a KV-cache. This cache stores the K and V vectors from previous computations, eliminating the need for redundant recalculation~\cite{agrawal_taming_2024}.

In inference scenarios of LLMs, there are two main processing stages. The first stage is known as the prefill phase. During this phase, the system processes the entire input sequence to produce the first output token, simultaneously populating the KV-cache with K and V vectors derived from the input. This stage primarily relies on matrix-matrix computations, which can be efficiently parallelized~\cite{agrawal_taming_2024}.

Following the prefill phase, new tokens are generated autoregressively,  leveraging the precomputed activations stored in the KV-cache. This is called the decode phase. The decode phase is computationally less demanding, often underutilizing the GPU \cite{agrawal_taming_2024}. To address this inefficiency, employing batching techniques becomes essential for optimizing LLM service, especially in high-load scenarios. 

There are two primary batching techniques: static and continuous batching, also called in-flight batching. Static batching aggregates all requests and processes them together until each has received a complete response. In contrast, continuous batching allows requests to dynamically join or leave the processing queue at each model iteration~\cite{Orca}. One iteration is the generation of a response token for all requests in the batch. We use in-flight batching for all models deployed on the Triton server to maximize throughput. 

Expert Router utilizes multiple independent models that are each processing their own batch of input requests. These multiple batches introduce another level of asynchrony in the inference system leading to distinct latency and throughput behaviours. 

\paragraph*{Triton Server}

We use NVIDIA's Triton server as an inference backend.
The Triton server starts with the inference process after receiving the prompt from the routing gateway (Step 5 in Figure~\ref{fig: Architecture}).  Each model is hosted on a dedicated Triton server, with each server assigned to its own GPU. This configuration allows the system to individually address models via distinct ports, each linked to a specific cluster identified by the k-means clustering algorithm. This architectural choice eliminates the need for communication between different models or GPUs, as each Triton server is equipped with its scheduling and in-flight batching mechanisms. We use each Triton server in streaming mode, i.\,e., tokens generated by the model are instantly transmitted back to the routing gateway, avoiding any buffering of tokens on the system side (Step 6 in Figure~\ref{fig: Architecture}). The routing gateway then forwards these tokens to the end-users through the NGINX load balancer (Steps 7/8 in Figure~\ref{fig: Architecture}). To enhance inference performance, we utilize TensorRT-LLM, a specialized framework designed to build and deploy LLM engines. 
\section{Experimental Setup}

This section describes the experimental setup designed to evaluate the impact of the Expert Router on inference performance. We describe the characteristics of the models integrated into the Expert Router and the reason behind their selection. Additionally, we present the baseline models identified through preliminary experiments aimed at optimizing performance. The section also details the User Simulator to generate high-load scenarios, enabling a thorough comparison between the Expert Router and the baseline models.

\subsection{User Simulator}
\label{subsec: user_sim}

The User Simulator manages N synthetic users, concurrently sending prompts to emulate increasing load on the inference systems. This is achieved by spawning N independent threads, each acting as a separate user. The simulated users draw their prompts from a selected segment of the C4 dataset \cite{raffel_exploring_2023}, which are distributed across eight categories defined by Gurungan et al.~\cite{Gururangan2023}. We conduct experiments using two different request scenarios to analyze the Expert Router configurations.

The first one is a uniform distribution across the eight categories, such that each model needs to process the same amount of queries. The rationale behind the uniform distribution is that it aligns with the scalability of the Expert Router. Such systems can dynamically and autonomously adjust by adding more models to balance spikes in demand for individual clusters, thereby maintaining uniformity in request distribution \cite{9284206}.

In the second scenario, we partition the requests across the eight categories using a normal probability distribution. This results in some models receiving more inference requests than others, leading to an unbalanced load that increases the dynamics within the system.  

\begin{figure}[htbp]
    \begin{center}
    \includegraphics{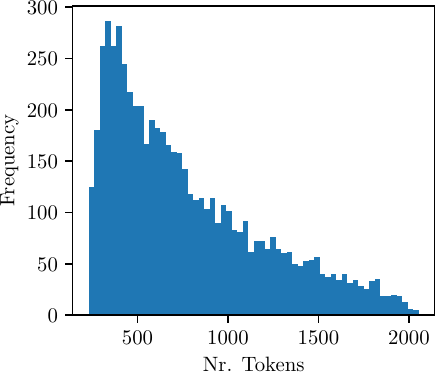}
    \caption{\textbf{Distribution of number of input tokens}. The plot shows the distribution of the number of input tokens in the test set. Its shape is based on the input length distribution used by MLPerf in the Llama 2 70B benchmark test \cite{atta-fosu_llama_2024}.}
    \label{fig:length_distr}
    \end{center}
\end{figure}

\begin{figure*}[ht!]
    \centering
    \includegraphics[height=6.5 cm]{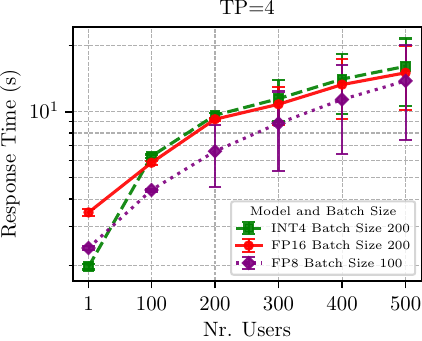}
    \hspace{0.5cm}
    \includegraphics[height=6.5 cm]{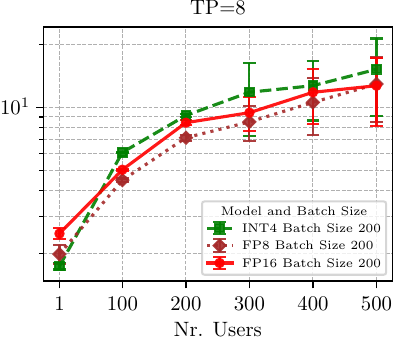}
    \caption{\textbf{Preliminary Experiments:} The two plots show the trajectories of the response times with increasing user concurrency for different batch sizes and data types. The results on the left side use a LLama 3 70B model parallelized over 4 GPUs (TP=4) and on the right side across 8 GPUs (TP=8). Based on these results we select the three baseline models listed in Table \ref{tab: Model_facts}.}
    \label{fig:Prelim}
\end{figure*}

The C4 dataset was selected due to its use by Gurungan et\,al.\ in developing their k-means clustering algorithm~\cite{Gururangan2023}. Each synthetic user generates a prompt, transmitted to the backend via gRPC. The distribution of input token lengths in our test dataset mirrors that of the dataset used by MLPerf in their Llama 2 70B inference benchmark test \cite{atta-fosu_llama_2024}. The token-length distribution across the test set is shown in Figure \ref{fig:length_distr}. This variability in input tokens introduces additional stress on the system, allowing us to investigate dynamic user behaviors more thoroughly.

Furthermore, we do not enforce a specific number of output tokens produced by the inference systems. The token generation is terminated with the arrival of the end of sequence token. However, we limit the maximum output length to 1000 tokens. Inference is carried out in streaming mode, allowing for measuring the arrival times of individual tokens. This data is stored and later used in the performance evaluation presented in Section \ref{subsec: eval}.

\subsection{Infrastructure Setup}

We use a DGX H100 for all experiments to host the baseline and expert router models. It contains eight H100 GPUs, each with 80 GB of GPU memory. To prevent interference between load generation and model inference, the User Simulator is deployed on a separate node. This node has an Intel Xeon Platinum 8480CL and two terabytes of RAM. The latency between the two nodes is measured at 17 $\pm$ 5 $\mu s$ and is negligible in the inference experiments. 

We deploy the Llama 3 70B model in the baseline configurations through a Triton server, employing tensor-parallelization across eight and four GPUs. The specific deployment details will be described in the next section. For the Expert Router experiments, the architecture hosts the routing gateway and individual Triton containers on the same GPUs. Specifically, we deploy 16 instances of the routing gateway, each requiring 1,138 MB of GPU memory, resulting in 2,276 MB on each GPU dedicated to the Expert Router system. The remaining GPU memory is allocated for deploying the individual expert models.

\subsection{Baseline Configurations}

To accurately benchmark and compare the performance of the Expert Router, we must establish well-optimized baseline models. Therefore, we conduct preliminary experiments to determine the best configuration by testing various batch sizes ($B_S=\{20, 100, 200, 400, 600\}$), data formats (FP16, FP8, INT4), and levels of tensor-parallelism (TP=4, TP=8). We evaluate each configuration by increasing the number of concurrent users from 1 to 500 in steps of 100.

Given the extensive combination of hyperparameters that are being tested, we limit the number of answer tokens to 200 to reduce the runtime of each experiment. Additionally, we use a test set with 335 $\pm$ 30 input tokens. Each test run is repeated five times to estimate variances in response times. All preliminary experiments are conducted using the Llama 3 70B model.

For TP=4 configurations, deploying two instances of the model is possible on a DGX H100. Therefore, each model runs on four dedicated GPUs. The preliminary experiments measure response times based on a single model parallelized across 4 GPUs. Two instances will be deployed for the final results presented in Section \ref{sec:results}, with requests distributed among the two models.

Fig~\ref{fig:Prelim} illustrates the response time trajectories as the number of users increases. The left plot shows the best-performing batch sizes for each data type under TP=4, while the right plot shows the results for TP=8.

For TP=8, the FP8 and FP16 models with a batch size 200 demonstrate similar performance when handling 500 concurrent users. However, for smaller numbers of users, the FP8 model outperforms the FP16 model. To ensure a comprehensive evaluation, both models will be included in the final experiments. We will refer to these two baseline models as A) 70B FP16 TP8 and B) 70B FP8 TP8 in the following sections. As shown in Table \ref{tab: Model_facts}, the reduced precision of baseline model B allows it to utilize over 100 GB more GPU memory for storing user context compared to baseline model A.

In the TP=4 results, the FP8 models consistently show a slight advantage. Therefore, in the final tests, we will deploy and evaluate two instances of the FP8 model with a batch size of 100 corresponding to baseline model C in Table \ref{tab: Model_facts}.

\subsection{Expert Router Configuration}

For the model configuration in the Expert Router, we utilize two different versions of the Llama3 model group: Llama3 70B and 8B. This approach allows us to examine how varying model specifications and configurations impact inference performance when integrated within the Expert Router. The first configuration (D) uses eight quantized 70B models to explore high-parameter scenarios; the second (E) involves eight billion parameters (8B) models, allowing for a larger KV-cache and FP16 weights. With every specification, our objective is to optimize the batch size within the constraints of available GPU memory, ensuring maximal utilization.

\begin{table}[h!]
\centering
\caption{\textbf{Model Specifications:} For all experiments conducted within this study, LLama 3 models are used. Besides the batch size, the table lists memory-related specifications for each of the models, such as the weight and the KV-cache size in GB.}
\renewcommand{\arraystretch}{1.5} 
\small 
\begin{tabular}{|l|c|c|c|}
\hline
\diagbox{Model}{Specs} & Weights & KV-cache & Batchsize \\
\hline
\bottomrule
\multicolumn{4}{|c|}{Baseline Models} \\
\toprule
\bottomrule
A) 70B FP16 TP8          &    148.8 GB      &   347.28 GB       &   200    \\
B) 70B FP8 TP8          &     69.6 GB      &      487.2 GB       &   200   \\
C) 70B FP8 TP4          &   69.6 GB      &     221.2 GB       &   100    \\
\hline
\bottomrule
\multicolumn{4}{|c|}{Expert Router Models} \\
\toprule
\hline
D) 70B INT4 (Exp.R.)         &    37.7   GB     &   37.6     GB      &    125    \\
\hline
E) 8B FP16 (Exp.R.)         &     15.4 GB     & 62.2 GB    &    125     \\
\hline
\end{tabular} 
\label{tab: Model_facts}
\end{table}

\paragraph*{D) 70B INT8 (Exp.R.)} The first configuration uses a Llama 3 70B model with INT4 quantized weights. We employ activation aware quantization (AWQ) \cite{lin_awq_2023} provided within the TensorRT-LLM library \cite{tensorrtLLM} to build the model. Table \ref{tab: Model_facts} shows that the resulting model uses 37.7 GB of GPU memory for its weights and allocates 37.6 GB of GPU memory for the KV-cache. The batch size is set to 125. Each GPU hosts a single INT4 70B model in this experiment, leading to 560 billion INT4 quantized weights across the system. In our tests, we included this particular model setup to explore how the Expert Router performs when handling models with many parameters.

\paragraph*{E) 8B FP16 (Exp.R.)} The second configuration uses Llama 3 8B model, requiring 15.4 GB for weight storage. This reduction in memory requirement for weights enables allocating a greater portion of available GPU memory to the KV-cache with 62.2 GB. This allows for evaluating  Expert Router's performance with non-quantized models that leverage a larger KV cache.

\section{Evaluation Methodology}
\label{subsec: eval}

Throughout our experiments, we assess the performance of the Expert Router and baseline models from two perspectives: user-centric and system-centric metrics. This strategy allows us to evaluate system-relevant metrics and analyze the effects of different Expert Router configurations on the interaction latency and throughput.

\subsection{User-Centric Metrics}

In the evaluation from the user perspective, we present three metrics. The first is the Time to First Token (TTFT) \cite{miao_towards_2023}. It measures the duration from when the user sends a request to when the user receives the first token. We report the TTFT average with an increasing number of users. Adjacent to the TTFT, we track the Time Per Output Token (TPOT). It describes the time difference between consecutive tokens. Again, we report the average TPOT for an increasing number of users. Finally, we also investigate the throughput from the user perspective. The user throughput measures the rate at which tokens are received, as experienced by a single user. This metric is calculated by dividing the total number of response tokens by the duration from the request's start ($t_s$) to the completion of the answer ($t_e$). We average this value across all users and report the mean and standard deviation.

\subsection{System-Centric Metrics}
\label{subsubsec: system-centric}
System-centric metrics evaluate the efficiency and responsiveness of the underlying infrastructure and software as they process tasks, especially when facing a rising number of concurrent user requests. To explore the impact of an increasing number of parallel user requests on system latency, we record the p99 response time for all responses. The p99 response time refers to the 99th percentile response time, meaning that $99\%$ of the requests are processed and responded to within this time frame. This allows us to investigate the effect of the Expert Router on the overall system latency.

\begin{figure*}[ht!]
    \centering
\includegraphics{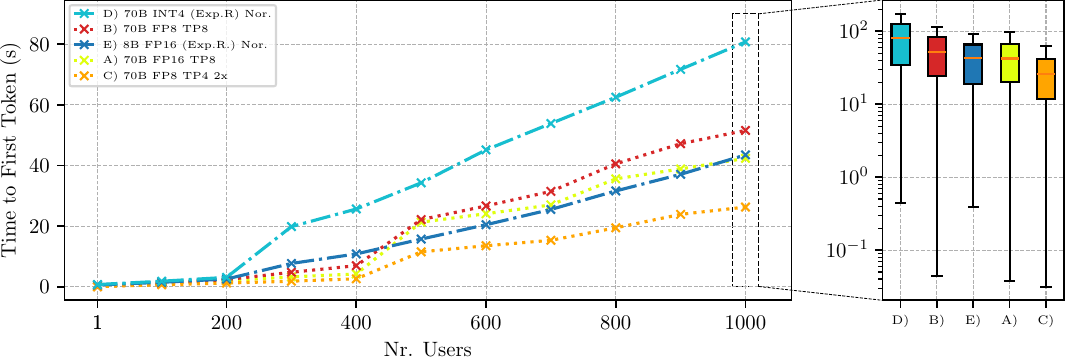}
    \caption{\textbf{Median Time to First Token:} Baseline models (A,B,C) show lower TTFT values compared to 70B Expert Router model (D) due to the benefit of tensor-parallelism in the prefill phase. The 8B Expert Router configuration (E) shows similar TTFT to tensor-parallized models due to reduced computational demands. The whiskers on the right box plot show that the Expert Router models have higher minimum values, highlighting the extra latency from the routing gateway.}
    \label{fig:TTFT}
\end{figure*}

In addition to traditional metrics, we use a method for evaluating the system throughput over time by analyzing the recorded token timestamps. In each trial, we establish K time intervals and count the number of tokens present within each interval. Thus, we only count how many tokens the system produces in a specific time interval, regardless of to which user a specific token belongs. This strategy for measuring the throughput allows us to concentrate on the distribution of token timestamps, enabling us to analyze the system throughput dynamics. To make the different systems more easily comparable we calculate the mean throughput across all time intervals and present these values for specific levels of user concurrency in Section \ref{subsec: ResultsSystem}.
\section{Results}
\label{sec:results}

In this section, we present the outcomes of our experiments, analyzing the characteristics of the Expert Router and baseline configurations from both user and system perspectives. We systematically increased the number of concurrent users sending requests to the system to assess the performance under varied load conditions.

In addition to the performance metrics presented in Section \ref{subsec: eval}, we have also measured the latency introduced by the routing gateway. The average latency for each user when sending prompts via the routing gateway is 442 ms with a standard deviation of 302 ms. This latency remains stable irrespective of the number of concurrent users. The high standard deviation is caused by outliers with very high latencies, which could originate from a bottleneck on the DGX system during peak load times. The latency of the routing gateway will be considered in the further discussion of the results. 
As presented in Section \ref{subsec: user_sim}, we will provide results for the Expert Router configurations under two different load scenarios: uniform and normal distributions. We will directly compare the two workload scenarios for the user throughput and the p99 response time. For all other metrics, we will focus on the results from the normal distribution, serving as an upper bound for latencies and a lower bound for throughput.

\subsection{User-Centric Metrics}

\begin{figure*}[hb!]
    \centering
\includegraphics{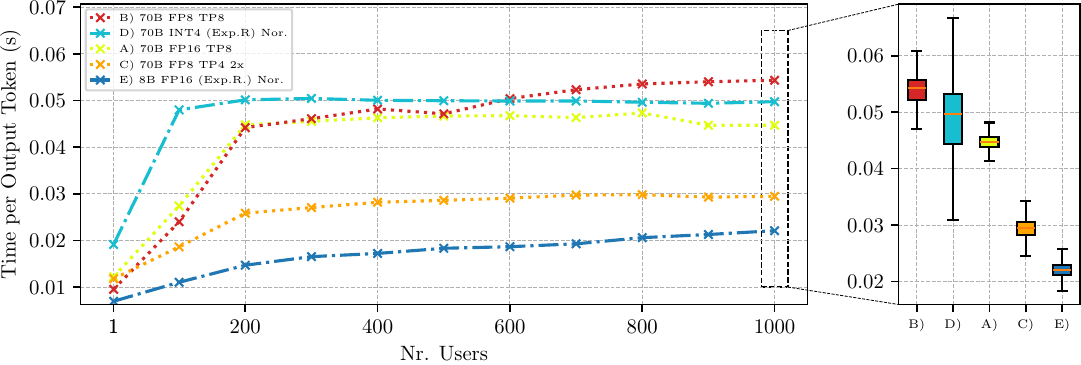}
    \caption{\textbf{Median Time per Output Token:} The 8B Expert Router configuration (E) maintains the lowest TPOT due to its smaller number of parameters. Baseline model C shows higher TPOT due to increased communication overhead with tensor parallelism.}
    \label{fig:TPOT}
\end{figure*}

\begin{figure*}[ht!]
    \centering
    \includegraphics{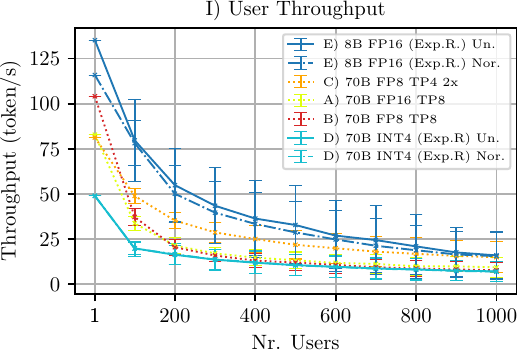}
    \label{fig:us_thr}
    \hspace{0.3 cm}
    \includegraphics{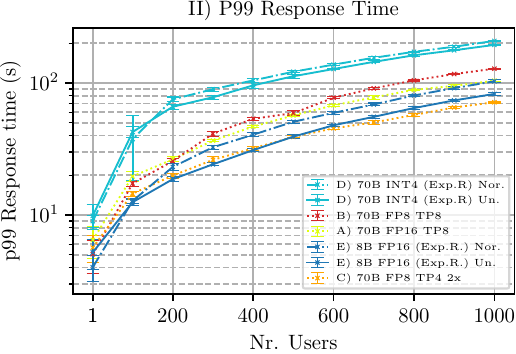}
    \caption{\textbf{User Throughput and p99 Response time:} Expert Router models (D, E) show a minimal difference between normal (Nor.) and uniform (Un.) workload distributions in both metrics, demonstrating robustness to varying workloads. In both metrics, model C and Expert Router configuration E show a clear advantage.}
    \label{fig:us_thr_p99_response}
\end{figure*}

\begin{figure*}[hb!]
    \centering
    \begin{minipage}[t]{0.33\textwidth}
        \centering
        \includegraphics{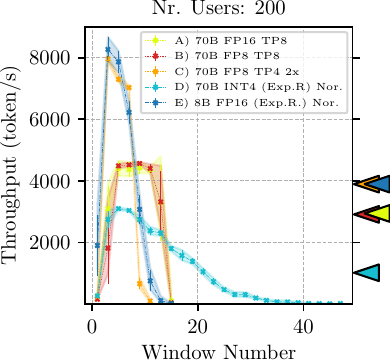}
    \end{minipage}%
    \hfill
    \begin{minipage}[t]{0.32\textwidth}
        \centering
        \hspace{0.57 cm}
        \includegraphics{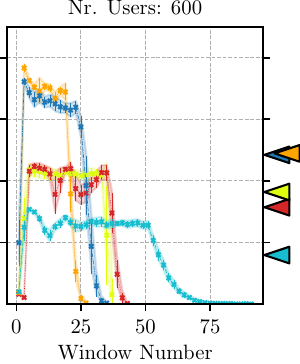}
    \end{minipage}%
    \hfill
    \begin{minipage}[t]{0.32\textwidth}
        \centering
        \includegraphics{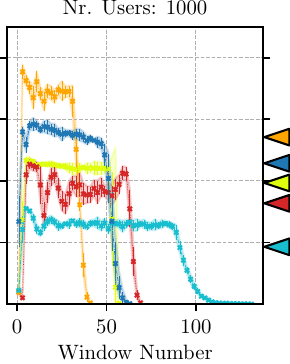}
    \end{minipage}
    \caption{\textbf{System throughput:} Expert Router configuration E matches the throughput of baseline model C at 200 and 600 concurrent users, but falls behind at 1,000 due to larger TTFT values. Configuration D consistently shows the lowest peak and average throughput, highlighting the trade-off of using high-parameter configurations. The colored arrows indicate average throughput. The window size in which the tokens are counted is set to 2s.}
    \label{fig:TimeBucket}
\end{figure*}

Figure~\ref{fig:TTFT} illustrates the Time to First Token (TTFT) for an increasing number of users. As visible in the box plot on the right, the baseline models (A, B, C) consistently achieve lower minimum TTFT values than the Expert Router models (D, E). This is caused by the additional latency introduced by the routing gateway in the Expert Router architecture. However, this added latency does not result in a significant bottleneck during inference, as visible by the constant relative difference in the median TTFT values in the left plot. Baseline model C, which uses two tensor-parallelized 70B models, consistently exhibits the lowest TTFT throughout all levels of concurrency.

The better performance of the tensor-parallelized baseline models (A, B, C) can be attributed to their ability to efficiently distribute computations during the prefill phase of inference across multiple GPUs. In this phase, the input's K and V vectors are computed and stored in the KV-cache. The parallel execution of the underlying matrix-matrix multiplications allows the tensor-parallelized baseline models to produce the first output token faster, leading to a lower overall TTFT~\cite{agrawal_taming_2024}. The 8B model requires fewer computations, leading to similar TTFT values for configuration E as the tensor-parallelized models.

The advantage of less computation for the smaller 8B Expert Router configuration (E) becomes even more visible in Figure \ref{fig:TPOT}. Configuration E shows a clear advantage in the Time per Output Token (TPOT) compared to all other models. This is further highlighted by the small variance of the TPOT for high levels of user concurrency, visible in the box plot on the right in Figure \ref{fig:TPOT}. The elevated TPOT for the baseline models is due to the increased communication overhead required for the tensor-parallelization.

The results suggest that for moderate levels of user concurrency, the number of parameters of the models in the Expert Router configuration does not drastically affect the TTFT. Furthermore, even with many concurrent users, the TPOT remains within acceptable ranges for both Expert Router configurations. Despite the differences in TTFT, these findings demonstrate that the Expert Router configurations can deliver reliable performance across varied workloads.

Figure~\ref{fig:us_thr_p99_response} I) evaluates the user throughput in response to increasing system load. This plot includes two result curves for each Expert Router configuration, representing requests distributed by normal (Nor.) and uniform (Un.) distributions across the clusters. For both Expert Router configurations, the difference between the two workload scenarios is minimal. This suggests that Expert Router configurations do not require perfectly balanced model utilization, and peaks in user requests have a limited effect on the user-perceived performance. Expert Router model E shows the highest user throughput up to a concurrency level of 800 users, highlighting its competitive performance from a user perspective. 

\subsection{System-Centric Metrics}
\label{subsec: ResultsSystem}

In the following, we assess the impact of increasing parallel user requests on the latency measured by the p99 response time and the system throughput. The suggested resilience of the Expert Router configurations to non-balanced workloads is also visible in the p99 response time trajectories shown in Figure \ref{fig:us_thr_p99_response} II). Both model configurations (D, E) only show minor differences between the normal and uniform request distributions. 
Baseline model C has the lowest p99 response time for high levels of user concurrency, caused by the previously shown stable TTFT.

To further assess the impact of Expert Router on system metrics, we present the system throughput shown in Figure \ref{fig:TimeBucket}. The size of one window in which the number of received tokens is counted is set to two seconds. For 200 and 600 concurrent users, Expert Router configuration E exhibits an almost identical throughput trajectory to baseline model C. In these scenarios, the Expert Router configuration does not negatively impact system throughput. This is further supported by the average throughput, indicated by the colored arrows on the right y-axis. It shows that both configurations achieve nearly the same average system throughput for these user counts. However, at 1,000 concurrent users, model C outperforms Expert Router configuration E. This highlights the effect of the larger TTFT values and the benefits of tensor parallelization on system throughput. 

The effect of using the high parameter 70B models in the Expert Router configuration is especially visible in these plots. Across all user numbers, model D has the lowest peak throughput. Although the average throughput increases with higher user concurrency, it does not pass the 2,000 tokens/s mark. These results highlight the trade-offs involved in deploying different model configurations. When the capabilities of multiple high-parameter models, like the 70B configuration, are required, user and system metrics are more significantly affected. The less parameter-intensive configuration can achieve a similar throughput to a tensor-parallelized model. It generally does not negatively impact user and system metrics across a wide range of concurrency levels.

\section{Conclusion and Future Work}
\label{sec:conclusion}

In this paper, we have characterized the performance of the Expert Router, a scalable routing architecture for LLM inference. The results demonstrate that the Expert Router effectively combines specialized models while maintaining robust performance in terms of throughput and latency, even with high concurrency levels and workload imbalances. Our characterization further shows that the latency introduced by the routing process remains constant even under high user concurrency, allowing model configuration, particularly the choice and tuning of parameters, to drive performance outcomes more significantly. For moderate levels of user concurrency, the inference metrics are not significantly impacted by the model size. In such scenarios, quantized high parameter models can efficiently increase the total parameter count without substantially compromising performance. At higher levels of user concurrency, smaller models can achieve very similar performance to tensor-parallelized baseline models. Such configurations require less advanced infrastructure and are especially well suited in computing environments without high-bandwidth inter-GPU connections. Conclusively, Expert Router provides a versatile solution for scalable LLM inference, adapting well to various compute environments and request scenarios while offering robust performance regardless of infrastructure limitations.

As part of future work, we plan to create a fully unsupervised pipeline for expert creation and inference. One main focus of this approach will be to conduct a detailed analysis of the model's ability to generate language, focusing on how different training datasets affect the performance of both the classifier and the expert models. Furthermore, we plan to extend this to dynamic deployment of LoRA weights, allowing us to increase the number of experts in the systems without increasing the number of models. Another avenue for future exploration is to study if different setups for the routing gateway can further increase the performance. Such experiments include testing different embedding and classification algorithms or expanding clustering criteria beyond the prompt domain to incorporate user and task categories.

\section*{Acknowledgment}
The authors generated parts of this text using OpenAI's language-generation models. Upon generation, the authors reviewed, edited, and revised the language.

\bibliographystyle{IEEEtran}
\bibliography{bib}
\vspace{12pt}
\color{red}

\end{document}